\documentclass[10pt,twocolumn,letterpaper,nofootinbib]{article}

\usepackage{wacv}
\usepackage{times}
\usepackage{epsfig}
\usepackage{graphicx}
\usepackage{amsmath}
\usepackage{amssymb}
\usepackage{comment}

\usepackage{xcolor}

\usepackage{cellspace}
\usepackage{caption,booktabs}
\usepackage[font=small,skip=0pt]{caption}
\usepackage{tablefootnote}
\usepackage{units}
\usepackage{xfrac}
\wacvfinalcopy 

\ifwacvfinal
\def\assignedStartPage{1}
\fi

\ifwacvfinal
\usepackage[breaklinks=true,bookmarks=false]{hyperref}
\else
\usepackage[pagebackref=true,breaklinks=true,colorlinks,bookmarks=false]{hyperref}
\fi

\ifwacvfinal
\else
\fi

\begin{document}

\title{PETA: Photo Albums Event Recognition using Transformers Attention}

\author{Tamar Glaser, Emanuel Ben-Baruch, Gilad Sharir, Nadav Zamir, Asaf Noy, Lihi Zelnik-Manor\\
\vspace{-2mm}
\\
DAMO Academy, Alibaba Group\\

{\tt\small { \{tamar.glaser, emanuel.benbaruch, gilad.sharir, nadav.zamir, asaf.noy, lihi.zelnik\} }}\\
\tt\small {@alibaba-inc.com}
}

\maketitle
\begin{abstract}
In recent years the amounts of personal photos captured increased significantly, giving rise to new challenges in multi-image understanding and high-level image understanding. Event recognition in personal photo albums presents one challenging scenario where life events are recognized from a disordered collection of images, including both relevant and irrelevant images. Event recognition in images also presents the challenge of high-level image understanding, as opposed to low-level image object classification.
In absence of methods to analyze multiple inputs, previous methods adopted temporal mechanisms, including various forms of recurrent neural networks. 
However, their effective temporal window is local. In addition, they are not a natural choice given the disordered characteristic of photo albums. We address this gap with a tailor-made solution, combining the power of CNNs for image representation and transformers for album representation to perform global reasoning on image collection, offering a practical and efficient solution for photo albums event recognition. Our solution reaches state-of-the-art results on 3 prominent benchmarks, achieving above 90\% mAP on all datasets. We further explore the related image-importance task in event recognition, demonstrating how the learned attentions correlate with the human-annotated importance for this subjective task, thus opening the door for new applications.\footnote{Code will be available at https://github.com/Alibaba-MIIL/PETA.}

\end{abstract}

\section{Introduction}
With the exponentially rising amount of personal photos, so rises the need for automated organization of photo albums. Manual album organization is no longer realistic for many people. Detecting valuable events in personal galleries and classifying their contents are essential core blocks for enabling such automatic organization. Event recognition is the process of classifying a specific collection of photos from a typical personal album into a predefined set of special events (e.g. Birthday, Trip, Sports etc). When dealing with photo album event recognition, we usually face three main challenges: 
(1) Analysis of multiple images disordered collection, with varying size and varying time gaps that range from seconds to hours.
(2) Typical photo collection usually includes many irrelevant photos. 
(3) Image representation should capture both high and low level information, however the available datasets are relatively small and might be insufficient for representation learning.

Previous approaches suggested several solutions, each deals only with part of these challenges. 
Multi-images analysis was addressed using simple or weighted averaging \cite{guo2015event, savchenko2020event}. In \cite{Wang_17_BMVC, qi2020discriminative, choi2020meta,guo2020graph} recurrent neural networks (RNN) were used. 
However RNNs are limited in capturing global relations between distant sequence items that is essential for photo albums event recognition.
Leveraging the more relevant album images was addressed through image importance ground-truth usage \cite{Wang_16_CVPR, Wang_17_BMVC, choi2020meta} however this requires exhaustive annotation of per-image importance.
Guo et al. \cite{guo2017multigranular} avoid this annotation and implicitly learn image importance through an attention layer comparing images to the event label. However, the implicit predicted image importance was not evaluated. To address the image representation challenge they further suggested an ensemble of several networks and hierarchical feature extraction, which may add complexity to the solution.
Savchenko \cite{savchenko2020event} also used an attention layer to aggregate album images feature descriptors with learnable image weights, yet their image representation includes a generative image captioning generation module which reduces efficiency and produces inferior recognition results.

In this paper we propose a practical and efficient transformers-based solution 
to deal with those challenges altogether and present state-of-the-art (SotA) results on three prominent event recognition benchmarks. \\
Transformers \cite{vaswani2017attention} have been growing in popularity as classification models for sequential data. 
Transformers enable multi-image analysis, while leveraging the more relevant items through attention mechanism.
While theoretically LSTMs \cite{hochreiter1997long} aims for both short-term and long-term memory, in practice information propagation is limited, as the furthest gradients effect may diminish. 
Transformers had made a breakthrough in processing sequential data, since the effective processing window limitations were removed. 
Through its parallelized computation scheme, the transformers architecture manages to consider the relations between distant sequence items and effectively applies global attention over the whole sequence. 
To address the high-level image representation challenge, while training using small datasets, we use knowledge transfer from a feature extractor trained on a general image classification task.
We show that our method can be used as a drop-in component with existing image-classification backbones, keeping its parameters fixed while training only the transformer component. 
In typical real-world applications of automatic photos organization, both single image classification and album event recognition are required. 
Image representation sharing between these two tasks could be a practical advantage, such as using pre-computed image representations.
Our approach allows both the knowledge transfer from large datasets, as well as parameters and computation sharing, similarly to multi-task-learning advantages. 

A core key in automatic album organization is the ability to identify which images contain valuable information, and differentiate those from the less relevant ones. This semantic content analysis task is beneficial for event recognition enhancement, but may be also useful for other image content ranking applications such as album summarization or filtering clutter images.
A major challenge in image importance prediction is its subjectiveness, therefore annotated datasets are merely available.
Previous approaches \cite{Wang_16_CVPR, Wang_17_BMVC, choi2020meta} trained models to directly predict image importance using explicit annotated data. 
Since Image importance is ill-defined, relying on human-annotated data is limited and may not scale easily due to the exhaustive annotation process. 
In our solution we propose to exploit the attention learned by the transformer as a prediction for album image importance.
We demonstrate our model's ability to predict image importance without being directly trained on this task-specific annotation. \\

Our key contributions are as follows:
\begin{enumerate}
    \item 
    A novel transformer-based approach for album event recognition which outperforms the known SotA significantly on three prominent benchmarks. 
    Our solution is efficient yet simple, leveraging pre-trained image feature extractor. For example, on ML-CUFED dataset\cite{Wang_17_BMVC} we achieve 90.09\% mAP, improving previous SotA by 7\%.
    \item 
    We further demonstrate our solution's strength for image importance prediction, implicitly learned by the transformer's attentions.
    We analyse our predictor by measuring its correlation with the human-annotated image importance, and show how it enables high quality image importance prediction with no explicit annotation or training.
    
\end{enumerate}

\section{Related Work}

Generally, event recognition could refer to several media types - single images (e.g. social media), personal photo collections, video or audio,  \cite{ahmad2019deep} reviews all 4 aspects. Many previous works \cite{ahmad2018ensemble, ahsan2017complex, laib2019probabilistic, guo2020graph} focus on the task of per-image event recognition. Per-image event recognition deals with the image representation challenge, but neither with multi-image analysis nor with irrelevant images.
Datasets like \cite{li2007and, ahsan2017complex, xiong2015wider} and even recent \cite{muller2021ontology}, are all generated 
by gathering unrelated images from different sources, and labeling each image separately with its event type. 
However, the three publicly available personal albums datasets - Holidays \cite{Holidays}, PEC \cite{bossard2013event} and ML-CUFED \cite{Wang_16_CVPR, Wang_17_BMVC} are all gathered from personal photo albums. They consist of both relevant and irrelevant images in each album, illustrating a more realistic scenario of personal user albums. 
While some previous works explored both single-image and album event recognition \cite{savchenko2020event} generally these are treated as two different tasks. 
While we focus on personal photos collections event recognition, as personal photo applications become more common, many relevant insights could be also adopted from single-image event classification on the image representation aspect. 

Event recognition in personal photo albums was previously explored by different approaches, combining between the single image representation challenge and the album representation challenge.
To address the image representation and small dataset size challenges
, Guo \cite{guo2015event} propose the use of an ensemble of networks. Each network is pretrained by different big dataset: a coarse network pre-trained on the large places-365 scene classification dataset \cite{zhou2017places}, then a fine network pre-trained on the large object-classification dataset ImageNet \cite{deng2009imagenet}.
They demonstrate that when trying to recognize an event, both low-level information such as objects as well as high level content information such as scenes are required. In their following work \cite{guo2017multigranular} expand these insights even further, adding an additional network to the ensemble. 
They also presented the use of attention layers - for each of the 3 CNN used, they aggregated the album images features using weighted average. The weights are learned using an auxiliary loss, that examines the similarity between the image feature representation, to the event label representation, using word2vec \cite{rehurek2010software} conversion. Hence they explicitly learn the image importance and show the improvement using this addition. Savchenko \cite{savchenko2020event} also used an ensemble of two networks, when one generates image caption from image embeddings. \\
For album images aggregation previous approaches used averaging \cite{guo2015event}, attention weighted averaging \cite{savchenko2020event}, RNNs, LSTMs \cite{Wang_17_BMVC, qi2020discriminative, choi2020meta} and GRUs \cite{guo2020graph}. These recurrent neural networks have, in practice, a limited temporal window. \\
Transformer made a breakthrough in NLP \cite{vaswani2017attention} by applying global attention over the whole input sequence.
Inspired by the recent transformers usage for videos \cite{sharir2021image}, 
expanding the use of transformers to image collections seems natural. 

In addition to the event recognition task, some related work deals with image importance prediction. \cite{Wang_16_CVPR} broadly discusses prediction of the image importance given the album event type, using a Siamese network and a designated ranking loss. Later \cite{Wang_17_BMVC} expended it to an iterative procedure of alternating between prediction of image importance and event type, and \cite{choi2020meta} further dealt with image importance prediction for for all event types. Both \cite{Wang_17_BMVC} and \cite{choi2020meta} also show the contribution to event recognition using the image importance prediction. These approaches all train the prediction network in a supervised manner, using the CUFED dataset annotated image importance. \cite{Wang_16_CVPR} broadly describes the high efforts required to annotate such data, for this subjective task. Expending the annotations automatically is one of \cite{choi2020meta} main contributions. 
Therefore, one can derive that predicting image importance in an unsupervised manner 
could be very helpful, and might even apply for other subjective image-content-based tasks.
We show that predicting image importance without explicitly training on its annotation achieves useful results.

\section{Event recognition using transformers}

\begin{figure*}
\begin{center}
\includegraphics[width=15cm]{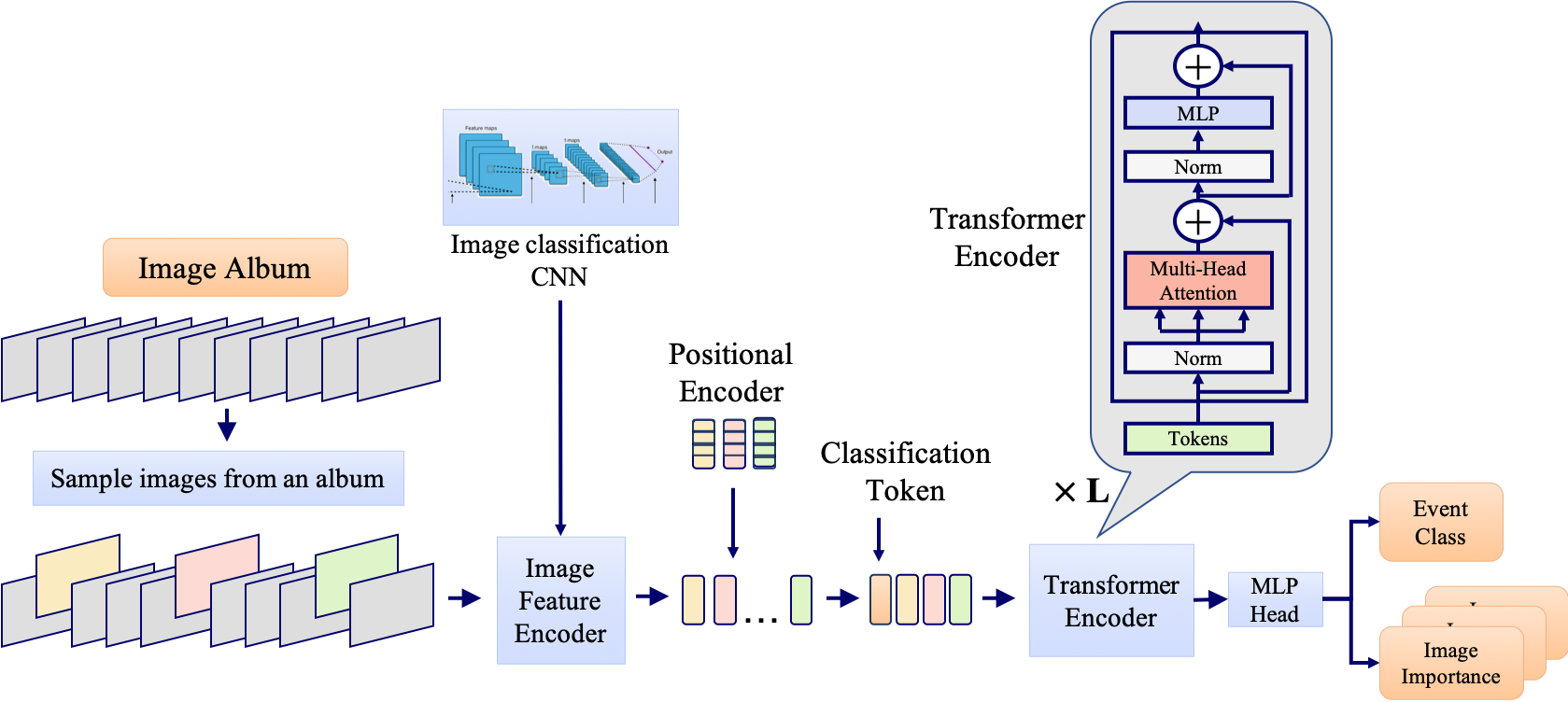}
\end{center}
   \caption{Our overall architecture.}
\label{fig:architecture}
\end{figure*}

Given a photo album with $N_A$ images: $\{I_t\}_{t=1}^{N_A}$, we would like to predict the event type $c\in \{1, ..., N_{cls} \}$ of the album. 
There are three main challenges in albums event recognition. First, album is a disordered variable length $N_A$ set of images. Second, albums consist of many irrelevant or partially relevant images. Third, high level image representation is required while albums event recognition datasets are rather small.
Computation efficiency and parameters constraints that every real-life system has, constitutes another challenge that should be considered. Our solution aims to achieve high recognition rates while dealing with all of these challenges.

Inspired by the recent success of transformers in many fields, both for NLP \cite{vaswani2017attention}, vision \cite{dosovitskiy2020image} and recently also in video \cite{sharir2021image, arnab2021vivit}, 
we aim to extend the use of transformers architecture to albums (image collections). To apply transformers for an image collection several adaptations should be made. 
Since album length is variable we sample a fixed number of images from an album $\{I_s\}_{s=1}^{S_A}$. \\
Next, we compute a representation vector (embedding) for each sampled image by applying a CNN model that was pre-trained on a large image-classification dataset.
These embeddings are then aggregated using transformers, which in turn, output the prediction layer for the event-type classification. In addition, an image importance prediction is extracted from the attention block. Image importance prediction is implicitly learned through the transformers training.
The different stages of our method are described next. Figure \ref{fig:architecture} illustrates the overall architecture of our proposed solution.

\subsection{Album Image Sampling}
\label{section:sampling}
As aforementioned, we aim towards a practical efficient solution. 
Since transformers are known to have a quadratic dependence on the length of the input sequence, we reduce the complexity by sampling a small number of images per album. 
Thanks to the attention mechanism enhancing
the more relevant photos, a smaller number of images can be used, while achieving high recognition rates. Images sampling allows both efficiency and dealing with the variable album length. 
We define a fixed number of sampled images per album $S_A$ as the number of images used for classification in both training and inference. 
Given an album of $N_A$ images $\{I_t\}_{t=1}^{N_A}$, we randomly sample $S_A$ images from the album.
We sample each batch $B$ by sampling $S_A$ images from $B_A$ albums:
\begin{equation}
    \nonumber B = B_A\cdot S_A
\end{equation}
In case of a short album with $N_A<S_A$, we pad to $S_A$ images by duplicating $S_A-N_A$ images. 

Analysis of the recognition rates and inference times for different selections of $S_A$ is described in section \ref{section:ipa}. We chose to use random permutation sampling as an album augmentation technique since time gaps between album images vary and images order isn't necessarily significant for event recognition. Image sampling is further justified through experimental results in section \ref{section:pos_embed_sample_order}.

\subsection{Image Representation}
\label{section:image_representation}
For the sampled images $\{I_s\}_{s=1}^{S_A}$, we apply an image classification backbone feature extractor:  $x_s = \text{CNN}(I_s)$
Event recognition datasets might not be large enough to train a sufficient image representation backbone. Therefore, we use an image classification pre-trained backbone to extract the sampled album images embeddings. This approach allows the backbone network to be used in a multi-task fashion for image classification in addition to album event recognition, both for knowledge transfer and parameters sharing.
Moreover, it allows offline image embeddings aggregation using only the transformers network, for the case of pre-computed image embeddings. 
The knowledge data transfer is demonstrated in section \ref{section:freeze}. 
Image representation encoder is a MTResNet \cite{ridnik2021tresnet} backbone, trained on Open Images V6 dataset \cite{OpenImages}. MTResNet is an efficient powerfull backbone, as demonstrated in section \ref{section:backbone}. Open Images V6 is a large varied image classification dataset consists of both low level and high level classes, hence prevents the need to combine networks trained separately on low level and high level classes, as done by \cite{guo2015event, guo2017multigranular}. Ablation on pre-train dataset selection are demonstrated in section \ref{section:pretrain_dataset}.

\subsection{Transformers Architecture}
\label{section:transformers_architecture}
The transformers architecture for image embedding aggregation is inspired by the temporal part of video frames aggregation technique used by STAM video action classification \cite{sharir2021image}
in which the sampled images are treated as tokens, and inserted into the transformer.
For each image embeddings $x_s$ a learnable positional encoding is added:
\begin{equation}
    \nonumber z^{(0)}_s = x_s + e^{pos}_s  , s \in \{1,2,..., S_A\}
\end{equation}
The positional encoding size is equal to the image embeddings dimension, in our case $D = 2048$. This is done for each image, generating a $S_A\times D$ matrix of images embeddings, for an album. 
To apply classification using transformers, a classification token $\text{CLS}$ \cite{devlin2018bert} is concatenated to the album representation,
expanding the album representation to $(S_A +1)\times D$. 
The classification token is a trainable vector, relying on the more relevant items \cite{devlin2018bert}, with the dimension as the input image tokens. The $(S_A +1)\times D$ album representation is inserted to stacked $L$ transformers encoder layers. The first encoder layer input is 
$ \left( \text{CLS}, \{z_s^{(0)}\}_{s=1}^{S_A}   \right)$. 
Since a multi-head self-attention layer with sufficient number of heads can directly model both short and long-distances interactions \cite{ramachandran2019stand}, each encoder consists of a Multi-head Self-Attention (MSA), with $H$ heads. 
For each encoder layer $l\in\{1,2,...,L\}$, and for each attention head $h\in \{1,2,...,H\}$, embeddings are normalized by LayerNorm (LN) \cite{ba2016layer} and linearly projected by the learneable $W_K^{(l,h)}, W_Q^{(l,h)}, W_V^{(l,h)}$ weights matrices, into Query, Key, and Value $Q,K,V$ for each input:
\begin{equation}
\nonumber Q_s^{(l,h)} = W_Q^{(l,h)} \text{LN}(z_s^{(l-1)})
\end{equation}
\begin{equation}
\nonumber K_s^{(l,h)} = W_K^{(l,h)} \text{LN}(z_s^{(l-1)})
\end{equation}
\begin{equation}
\nonumber V_s^{(l,h)} = W_V^{(l,h)} \text{LN}(z_s^{(l-1)})
\end{equation}
Then, using a scaled-dot-product attention the weighted values are computed, scaling by attention head dimensions 
$D_H=\nicefrac{D}{H}$ :
\vspace{-4mm}
\begin{equation}
z(Q,K,V) = \text{softmax}\left( \frac{QK^T}{\sqrt{D_H}} \right)V
\label{eq:softmax}
\end{equation}
Attention heads output, for each layer $l$, are concatenated and passed through a 2 Multi-Layer Perceptron (MLP) layers with GeLU \cite{hendrycks2016gaussian} activations. 
LayerNorm \cite{ba2016layer} is applied before MLP, and both MSA and MLP layers are operating as residual operators, due to the skip connections added.
The output of the network is the multi-label prediction, and an attention vector relating each image token to the classification token. 
This output is further explored in section \ref{section:attn_interp}.
The main benefit achieved by the transformers architecture is the attention-based album images aggregation, which leverages the more relevant images and reduces the low-relevancy images contribution. The weighted value extracted from the multi head attention could be referred to as the different images weights. 
Relevant images emphasis by attention mechanism allows good recognition with a small number of randomly sampled images.
In our implementation, we chose $L=6$ transformers encoder layers, and $H=8$ heads for the multi-head-self-attention module.

\section{Image Importance Prediction}
\label{section:attn_interp}

The ability to focus on the more relevant album images is a key component in album event recognition. Both \cite{Wang_17_BMVC} and \cite{choi2020meta} showed that image importance prediction improves event recognition results. However all previous approaches predicted image importance using annotated ground-truth while training \cite{Wang_16_CVPR,Wang_17_BMVC, choi2020meta}. This subjective task annotation is highly complex. Recently \cite{choi2020meta} proposed to automatically expand it for all event classes, stating the difficulty of this subjective annotation process.

The attention mechanism enables capturing of the global relations between the images, which in turn, may imply which are more relevant for the classification. 
In this section, encouraged by the event recognition results, we explore the ability to predict image importance from the learned attention, without using task-specific annotation.

From the last transformers encoder $L$, we extract the scaled-dot-product as in equation \ref{eq:softmax}. 
The output is an attention matrix:
\begin{equation}
    \nonumber
    z^{(L,h)}(Q,K,V) \in \mathbb{R}^{(S_A+1) \times (S_A+1)}
\end{equation}
Where $a\in \{1,..,H\} $ is the head number of the multi-head attention. We average the attention output over all heads:
\begin{equation}
    \nonumber
    A^{(L)} = \frac{1}{H}\sum_{h=1}^H \left[ z^{(L,h)}(Q,K,V)\right]
\end{equation}
Recall that the classification token was concatenated to the transformer's input as the first token, we extract image importance for image $I_s$ from the first row of the attention matrix.

Let us define $\gamma_s$ as the prediction of image importance for image $I_s$. 
We can simply extract the image importance by
\begin{equation}
\nonumber
\gamma_s = A^{(L)}_{0,s+1} \quad ,\quad  s \in \{1,..., S_A\}
\end{equation}
Where $ A^{(L)}_{i,j} $ is the j\textit{th} element in the i\textit{th} row of the attention matrix $ A^{(L)} $. 

This is not accurately the pure image importance for image $I_s$, as it involves other images with lower weights throughout the $L$ encoder layers. However it does reflect the importance of image $I_s$ within the album.
We performed some quantitative experiments in order to examine the learned attention predictor, by comparing it to the image importance human annotated ground truth provided with the ML-CUFED dataset, these are described in section \ref{section:attn_prediction}. In addition, we qualitatively explore the album images ranking, to demonstrate this prediction significance. 

\section{Datasets}
As aforementioned, publicly available datasets for event recognition are limited. This constitutes challenges for photo album classification. Apart from their sizes, there are only three publicly available personal photo albums events datasets. There are a few additional per-image event recognition datasets, such as \cite{li2007and, ahsan2017complex, xiong2015wider} and recent \cite{muller2021ontology}. These address only the image representation challenge and not the multi-images or irrelevant images challenges. There are also some private datasets \cite{guo2017multigranular, lonn2019smartphone}. We unprecedentedly demonstrate our results on all three publicly available datasets. All consist of authentic personal albums, each of which includes many irrelevant images. These datasets are described next.

\subsection{ML-CUFED dataset}
The CUration of Flickr Events Dataset (CUFED) \cite{Wang_16_CVPR} is the most recent and largest albums event recognition dataset. ML-CUFED is its expansion to multi-label events \cite{Wang_17_BMVC}. It includes 94,798 images in 1,883 albums, with 30-100 images in each album. It has 23 event classes, including personal events such as Birthday, different trips events, holidays, sports, etc. 
In addition to the event type annotation, the ground-truth includes an additional annotation, called "image importance". This annotation was done by asking several annotators how relevant is each image to the event, and averaging the score over annotators. 
The mean Spearman correlation between annotators, computed by several 2-groups splits, is 0.4, which shows how subjective the task is. 
Following previous papers  \cite{Wang_17_BMVC, savchenko2020event} we also generated a 4:1 train-validation for training and evaluating our approach.

\subsection{PEC Dataset}
The Personal Event Collection - PEC \cite{bossard2013event} dataset contains 61,364 images in 807 albums. Each album is annotated with a single label out of 14 classes. Event classes include personal events such as Birthday, trips like Boat cruise, Hiking and holidays such as Christmas, Halloween. The split provided with the dataset was used in our experiment.
 
\subsection{Holidays dataset}
The Holidays dataset \cite{Holidays} includes albums for the top 10 most frequent American holidays according to Google's Picasa albums - Christmas, Halloween, Easter, etc. It consists of about 50 albums for each holiday, 565 albums and 46,609 photos in total. Data gathering is based on human generated albums with names that indicate a specific holiday. Therefore the data illustrates real personal user gallery albums, including many irrelevant images within each album. 

\begin{table*}[hbt!]
\centering 
\begin{tabular}{Sc Sc Sc Sc Sc Sc}
\hline \\[-1.5\medskipamount]
Method & GT & ML-CUFED & PEC &  PEC & Holidays \\ 
  &   & mAP  & mAP & Accuracy & mAP \\ 
\hline\\[-1.5\medskipamount]
Wang\cite{Wang_17_BMVC} & events-only & 82.9 & -- & 85.5 & -- \\
Wang\cite{Wang_17_BMVC} & events+importance & 84.5 & -- & 87.9 & -- \\
Guo\cite{guo2017multigranular} & events-only & -- & 90.07 &  -- & -- \\
Choi\cite{choi2020meta} & events+importance (extended) & 75.7\tablefootnote{On single-label CUFED dataset} & -- & 91.1\tablefootnote{No human-annotated image importance, automatically generated} & --\\
CNN-aggregated & events-only & 80.91 & 83.1 & & 79.66 \\ 
\textbf{PETA} & events-only & \textbf{90.09} & \textbf{96.26} &  \textbf{96.45} & \textbf{97.51}\\ [0.6ex]
\hline
\end{tabular}
\caption{Event recognition Results for all 3 datasets} 
\label{table:recognition} 
\vspace{-4mm}
\end{table*}

\section{Experimental Results}

Several experiments were conducted in order to both demonstrate the event recognition results and the attention learning. For each event recognition dataset we compare our results to the best reported results.
\subsection{Event Recognition Results}
\label{section:rec_results}
\begin{figure}[htbp!]
\begin{center}
\includegraphics[width=0.95\linewidth]{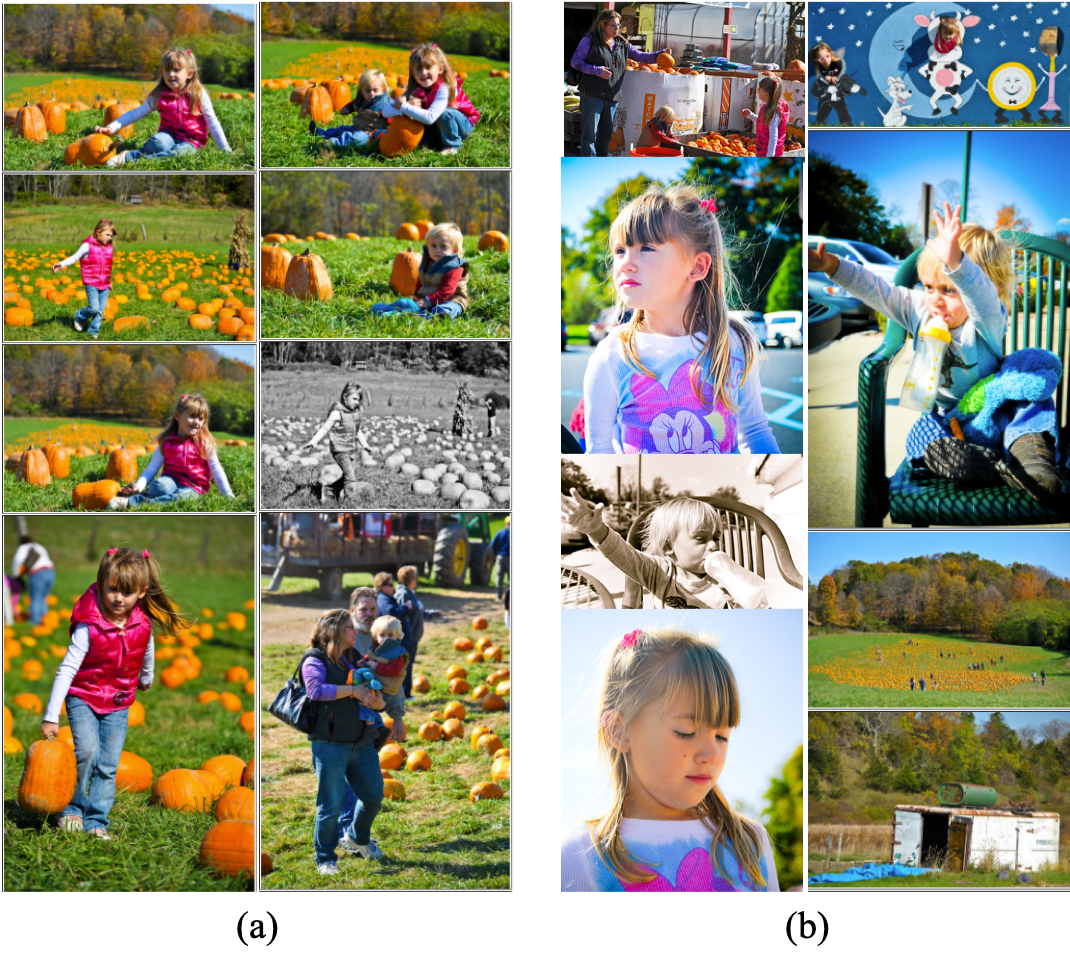}
\end{center}
   \caption{\textbf{Halloween event sample images}, from ML-CUFED dataset - recognized as Halloween by our approach, mistakenly recognized as Birthday by a CNN confidence averaging. (a) Top 8 attention images (b) Lowest 8 attention images.}
\label{fig:recognition}
\vspace{-4mm}
\end{figure}

We applied our event recognition network on the three publicly available datasets. For all datasets, we significantly outperform SotA. Recognition results are presented in table \ref{table:recognition}. 
All results are measured using mAP.
For PEC dataset we also measured mean accuracy, since some papers report mAP and some report accuracy.\\
Our results for ML-CUFED gets a mAP of \textbf{90.09\%}, which is 7\% improvement over the best method with no image importance ground-truth 
(Some reported results on ML-CUFED were trained with image importance ground-truth)
For PEC dataset we improve mAP by 6\% w.r.t. best reported mAP \cite{guo2017multigranular},
and we improve previous SotA mean accuracy by 5.3\% \cite{choi2020meta}.
Recent reported results on the holidays datasets are quite low and irrelevant therefore we compare to the CNN confidence aggregated baseline network, improving by $\sim18\%$.
The significant improvement achieved by the proposed method demonstrated the effectiveness of the transformers for album classification. Figure \ref{fig:recognition} shows some images sampled from a typical Halloween event in ML-CUFED dataset. Our approach correctly recognized a Halloween event, while simple per-image confidence predicted a Birthday event, although the former is done using only $S_A = 32$ randomly selected images, while the latter is done using all album images.
The prediction difference could be explained by examining the image importance values predicted. (a) Displays the top 8 attention images, while (b) displays the lowest 8 attention images. The higher attention images seem highly relevant, while the lower ranked images seem highly irrelevant.

\subsection{Image Importance Prediction Results}
\label{section:attn_prediction}

In this section, we examine the ability of our proposed approach to predict the importance of each image in the photo album. To measure it, we use the Spearman correlation metric between the image importance ground truth and the predicted values. Spearman correlation was selected as a metric since it allows global ranking measurement, considering both higher and lower ranked images.
We compared the results of three schemes: random ranking, per-image confidence of a convolutional network, and the learned attention using our architecture. Table 2 shows the the mean Spearman correlation computed for the different approaches. As can be seen, using the learned attention, the Spearman correlation value was significantly increased compared to the other baselines which indicates how informative and effective the learned attention is for a reliable estimation of the image importance in a personal album, without the need for explicit annotation.

\begin{figure*}[htbp!]
\begin{center}
\includegraphics[width=0.9\linewidth]{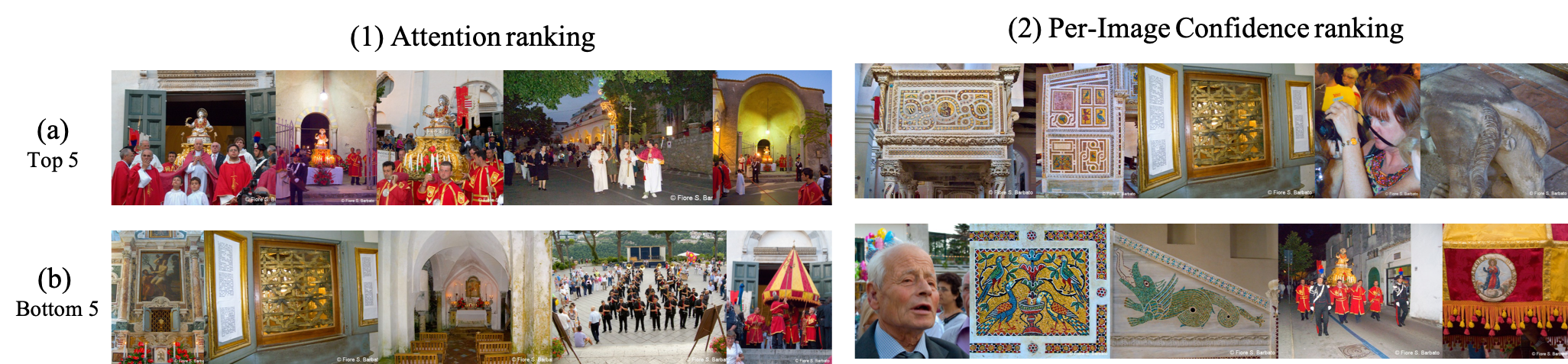}
\end{center}
   \caption{\textbf{Ranking by image importance.} For a "Religious activity" event from ML-CUFED, a qualitative comparison of the ranking by (1) Our proposed learned attention (2) per-image CNN confidence, when (a) top 5 ranked images, (b) bottom 5 ranked images.}
\label{fig:rank_religious}
\end{figure*}

\begin{table}[hbt!]
\centering
\begin{tabular}{Sc Sc} 
\hline
Predicted & Mean Spearman correlation \\ [0.5ex]
\hline
CNN Confidence & 0.3 \\
Transformers attention & 0.374 \\
Human annotators & 0.4 \\
\hline
\end{tabular}
\caption{Mean Spearman correlation over albums, ML-CUFED}
\label{table:corr-cufed}
\vspace{-2mm}
\end{table}
To qualitatively illustrate the ability of the transformers to capture the image importance in the album, Figure \ref{fig:rank_religious} shows an example of image ranking result using our image importance prediction and compare it to the ranking results obtained with the naive approach based on network’s confidences. As can be seen, for a Religious activity event from ML-CUFED dataset, (1a) top-5 ranked images by our predicted image importance seem highly relevant, while (1b) lower 5 ranked images seem less relevant. However when examining the ranking by the naive per-image prediction confidence (2a) top-5 images are less relevant, while some relevant images appear in (2b) the lowest ranked images.

\subsection{Ablation Experiments}
Our solution is composed of several modules aiming to combine the most efficient modules for each part of the architecture. To justify our selections, we compare different alternatives and examine their effect in an isolated manner. Most of our ablation studies experiments were performed on ML-CUFED dataset, as it is the largest and the most recent dataset. It is also  the only multi-label dataset, which is a more challenging and practical problem.
\subsubsection{Images per Album effect}
\label{section:ipa}
As presented in section \ref{section:sampling}, a subset of images are first sampled from the album and their representation vectors are fed into the transformer. In this section, we present the impact of the number of images per album ($S_A$) on the recognition results.
One of the main reasons for the transformers strength, as shown, is its ability to adaptively weight the different images, and increase the attention of the more relevant ones. This also allows the model to accurately classify event with less images. 
Table \ref{table:ipa-cufed} shows that even when using only $S_A = 8$ or $S_A = 16$, the results outperform previous approaches, which used all album images. In section \ref{section:rec_results} we compared our results using $S_A=32$ to previous benchmarks since the latter were using the whole album (e.g. an LSTM architecture does not require a fixed number of images). In practice, even using $S_A=32$ constitutes $\sim50\%$ or even $\sim30\%$ of the album images in many albums, therefore the comparison in this aspect isn't perfect as our architecture uses less images even for $S_A=32$. The transformers architecture does require a fixed number of images, but it's aggregation strength allows it to recognize the event type even with a small portion of the album's images. 

\begin{table}[htb!]
\centering 
\begin{tabular}{p{2.4cm} p{2cm} p{2.6cm}}
\hline
Images per album ($S_A$)  & mAP[\%] & Inference time per album [ms]\\ [0.5ex] 
\hline
4 & 81.8 &  3.01\\ [0.5ex]
8 & 86.89 &  5.86\\ [0.5ex]
16 & 88.58 & 12.06\\[0.5ex]
32 & 90.09 & 24.29 \\ [0.5ex]
\hline
\end{tabular}
\caption{Event recognition results for different number of sampled images per album, ML-CUFED dataset}
\label{table:ipa-cufed} 
\vspace{-6mm}
\end{table}
\subsubsection{Backbone Comparison}
\label{section:backbone}

In our architecture we used MTResNet \cite{ridnik2021tresnet} network for image representation. MTResNet is an efficient network that preserves the representation power with less parameters. Since the previous best reported results on ML-CUFED and PEC \cite{Wang_17_BMVC,choi2020meta} used ResNet-101 and ResNet-50 \cite{he2016deep} respectively, we also reported the results obtained with these backbones as well in Table \ref{table:backbone-cufed}. As shown, significant improvements were achieved on both ML-CUFED and PEC datasets for all the backbones, while MTResNet gives a good accuracy-efficiency trade-off.

\begin{table}[htbp!]
\centering 
\begin{tabular}{m{0.155\textwidth} m{0.065\textwidth} m{0.04\textwidth} m{0.04\textwidth} m{0.065\textwidth}}
\hline
Backbone & MLCUFED & \multicolumn{2}{Sc}{PEC} &\#Params \\ [0.5ex]
 & mAP & mAP & acc. &  \\ [0.5ex]
\hline 
\cite{choi2020meta} ResNet50 &  &  & 91.1 & 23M \\ [0.5ex]
PETA+ResNet50 & 89.13 & 94.63  & 91.66 & 23M \\[0.5ex]
\cite{Wang_17_BMVC} ResNet101 & 84.5\tablefootnote{With image-importance ground truth} & &  & 44.65M \\[0.5ex]
PETA+ResNet101 & 90.07 & 96.63 & 91.6 & 44.65M \\[0.5ex]
PETA+MTResNet & \textbf{90.09} & 96.26 & \textbf{96.45} & 29.4M\\ [0.5ex]
\hline
\end{tabular}
\caption{Event recognition results for different backbones}
\label{table:backbone-cufed} 
\vspace{-5.5mm}
\end{table}
\subsubsection{Backbone Freezing Comparison}
\label{section:freeze}

As described in section \ref{section:image_representation}
we applied transfer learning from an image classification network while freezing its parameters. In this section, we examine the impact of unfreezing the parameters of the image feature encoder network.
Table \ref{table:backbone-freeze} compares the mAP achieved with and without freezing the backbone’s parameters, showing that frozen backbone achieves similar or higher recognition rates. Therefore 
our hypothesis regarding 
knowledge transfer from a large dataset 
is confirmed,
in addition to the practical advantages of parameters and computation save.

\begin{table}[hbtp!]
\centering
\begin{tabular}{Sc Sc Sc}
\hline
  & ML-CUFED [mAP] & PEC [acc.] \\ [0.5ex]
\hline
Unfrozen Backbone & 89.31 & 94.05  \\ 
Frozen Backbone & 90.09 &  96.45 \\
\hline
\end{tabular}
\caption{Event recognition results with frozen vs. unfrozen backbones}
\label{table:backbone-freeze} 
\vspace{-7mm}
\end{table}

\subsubsection{Pre-Train Dataset}
\label{section:pretrain_dataset}
The need for transfer learning and the dataset used for pre-training was broadly discussed in previous works \cite{ahmad2016hierarchical}, \cite{guo2017multigranular}, \cite{ahsan2017complex} which even combined ensembles of networks pre-trained on different dataset types - object dataset, scenes datasets, and additional datasets. We chose to use Open Images V6 dataset \cite{OpenImages} as it includes 19,957 varied image-level classes, of both low-level objects and high-level descriptions. We compared our results pre-training on ImageNet  \cite{deng2009imagenet} and on ImageNet21K \cite{ridnik2021imagenet}. As can be expected, table \ref{table:pretrain-dataset} shows that using Image-Net-1K as a pre-train dataset indeed produces inferior results w.r.t. OpenImages, but it also demonstrates that this configuration benefits from the transformers architecture as well. 

\begin{table}[!htbp]
\centering 
\begin{tabular}{Sc Sc}
\hline
Pre-Train Dataset & mAP [\%] \\ [0.5ex]
\hline
ImageNet & 85.38 \\
ImageNet21K & 89.65 \\ 
OpenImagesV6 & 90.09 \\ 
\hline
\end{tabular}
\caption{Event recognition results for different pre-train datasets, ML-CUFED dataset}
\label{table:pretrain-dataset} 
\vspace{-7.3mm}
\end{table}

\subsubsection{Multi-label Loss}
\label{section:loss}
The ML-CUFED dataset \cite{Wang_17_BMVC} is a multi-label dataset with 23 classes. For training a multi-label classification model, we activated each output logit by an independent sigmoid function. One of the major challenges in multi-label classification problems is the high negative-positive imbalance. Therefore, we adopted the asymmetric loss (ASL)  proposed in \cite{ben2020asymmetric} to dynamically control the weight of the positive and negative loss terms.
In table \ref{table:loss}, we compared the mAP results obtained with three different loss functions for multi-label classification.  We found that the ASL is better suited for our task and used it in all our experiments.

\begin{table}[!htbp]
\centering 
\begin{tabular}{Sc Sc}
\hline
Loss & mAP [\%] \\
\hline
Cross-entropy & 89.39 \\
Focal-Loss \cite{tsung2017focal} & 88.17 \\ 
ASL\cite{ben2020asymmetric} & 90.09 \\ 
\hline
\end{tabular}
\caption{Event recognition results for different multi-label losses, ML-CUFED dataset}
\label{table:loss} 
\vspace{-5mm}
\end{table}

\subsubsection{Sampling and Positional Encoding}
\label{section:pos_embed_sample_order}
There are several alternatives for the image sampling required for our solution. For video frames sampling \cite{sharir2021image} the native choice is uniform ordered sampling. 
However in a personal album, the captured images frequency is not fixed and known as in video. 
The time gaps between photos could vary from seconds to hours. 
This significantly reduces the essence of the sequence order
as there's no temporal relation between the images in many cases. Therefore when sampling only a small portion of the event, random sampling could fit just as uniform sampling. To allow album augmentation while training we chose to sample with random permutations. 
Since the number of albums is rather small - 565, 807, 1887, simulating augmented albums has some generalization potential. The positional encoding could have a position significant if it is computed out of the original image timestamp. When timestamp is unavailable, positional encoding is used only for album augmentation. Table \ref{table:pos_embed_sample_order} shows examination of the different sampling approaches, and verifies our hypothesis.
\begin{table}[!htbp]
\centering 
\begin{tabular}{Sc Sc}
\hline
Sampling & mAP [\%] \\
\hline
Uniform ordered + pos. encoder & 88.59 \\
Random permutations  & 89.86\\ 
Random permutations + pos. encoder & 90.09 \\ 
\hline
\end{tabular}
\caption{Event recognition results for different images sampling methods, ML-CUFED dataset}
\label{table:pos_embed_sample_order} 
\vspace{-6mm}
\end{table}

\section{Conclusion}
Event recognition in personal photo albums is a high-level semantic task that incorporates both low level and high level image content understanding, as well as album image aggregation. We proposed a solution that combines efficiency and practical constraints, while significantly outperforming SotA recognition results. We demonstrated our superiority on all three publicly available datasets. We showed that transformers architecture enables dealing with a variable-length image collection with irrelevant images. In addition, we showed that the transformers can learn image importance prediction in an unsupervised manner. We believe that this approach can be used in practice in personal photos applications, as well as leverage subjective image content-based scoring without the need for specific annotation.

{\small
\bibliographystyle{ieee_fullname}
\bibliography{egbib}
}

\end{document}